\documentclass[letterpaper,10pt,conference]{ieeeconf}
\IEEEoverridecommandlockouts
\usepackage{amsmath,amssymb,graphicx,booktabs,float}

\usepackage{xcolor}
\definecolor{projectblue}{HTML}{586C94}
\usepackage[colorlinks=true, urlcolor=projectblue,
            linkcolor=black, citecolor=black]{hyperref}
\graphicspath{{figures/}}

\title{Learning Latent Goal Costs for Visual World-Model Planning with an Expanding Curriculum}

\author{
  \authorblockN{Konstantinos Mylonas \qquad Thrasyvoulos Spyropoulos}
  \authorblockA{
    Electrical and Computer Engineering, Technical University of Crete\\
    \texttt{kmylonas@tuc.gr} \quad \texttt{spyropoulos@tuc.gr}\\[0.6em]
    \href{https://super-block-2a9a.kmylonasece.workers.dev/}{Project website}
  }
}

\begin{document}
\maketitle

\begin{abstract}
Image-goal planning with a frozen visual world model depends on how predicted observations are scored against the goal. Mean-squared error (MSE) between embeddings from a pretrained visual encoder can provide a useful signal near a goal, yet favor states across an obstacle that are visually similar to the goal and difficult to reach. We fit a small residual adapter on top of frozen I-JEPA embeddings to improve this latent goal cost. Starting near each goal, a planner collects trajectories from progressively wider regions of initial states. The adapter learns from the order of observations along successful trajectories, with additional comparisons inferred from failed rollouts. No task-specific progress labels are used for fitting, so the procedure can be applied to other goal-reaching tasks. the DotWall and PointMaze used in this work. In paired 50-case evaluations with identical starts, goals, world models, and planner settings, success increases from 54\% to 98\% in DotWall and from 82\% to 96\% in PointMaze. On 16 unseen DotWall layout--goal cases, the adapter fitted on one fixed layout improves agreement between latent-cost ordering and shortest legal-path ordering from 71.9\% to 85.8\%. The results indicate that planning experience can improve a frozen visual representation's goal cost without retraining the encoder or the world model.
\end{abstract}

\section{Introduction}

World models have long been useful for decision making and control because they let an agent predict the consequences of its actions before acting \cite{hafner2019planet,hafner2020dreamer}. Visual latent world models make these predictions in a learned latent space: they predict the next latent state from previous latent states and actions. Once the model has learned the environment's dynamics, a planner can roll out candidate action sequences and score their predicted outcomes against a goal. In closed-loop model predictive control, the agent executes the first action of the selected plan, receives a new observation, and replans. Predicting in latent space avoids expensive pixel reconstructions and allows many action sequences to be evaluated efficiently.

\begin{figure}[b]
\centering
\begin{minipage}[t]{.36\columnwidth}
\centering
\includegraphics[width=\linewidth]{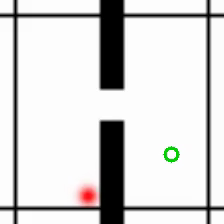}\\[-1mm]
{\small (a) DotWall}
\end{minipage}\hspace{.06\columnwidth}
\begin{minipage}[t]{.36\columnwidth}
\centering
\includegraphics[width=\linewidth]{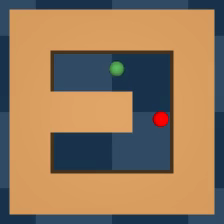}\\[-1mm]
{\small (b) PointMaze}
\end{minipage}
\caption{Rendered observations from the two environments. The red ball starts at its shown position and must reach the green goal. In DotWall, the legal route passes through a doorway; in PointMaze, it goes around an obstacle.}
\label{fig:environments}
\end{figure}

Planning this way still presents two challenges. Errors accumulate when a world model is rolled out for many steps, limiting how far ahead it can predict reliably. Separately, the latent cost between a predicted state and the goal need not reflect how difficult it is to reach the goal from that state. This is particularly relevant when a world model uses a pretrained, task-agnostic visual encoder, as in DINO-WM, which uses DINOv2 to encode image frames \cite{zhou2025dinowm, oquab2024dinov2}. The resulting embeddings can capture visual similarity without encoding the travel required by a particular task. Consider DotWall in Fig.~\ref{fig:environments}. The ball can appear close to the goal across the wall, even though it has to travel through the doorway to reach the other room. A planner using raw latent cost may therefore favor a state that looks promising, although in reality it is difficult to reach the goal from it \cite{alrasheed2026planninglimitslatentworld}. 

We study this problem with a world model built on a pretrained I-JEPA encoder \cite{assran2023ijepa}. Pretrained visual encoders have useful spatial and geometric features that extend beyond one navigation task \cite{assran2023ijepa,oquab2024dinov2}. We keep the I-JEPA encoder and world model fixed, and fit a small adapter to the encoder's latent representations so that their distance to the goal guides planning better than the raw latent cost. In DotWall and PointMaze, we know what an ideal measure of progress looks like: geodesic distance decreases along a shortest legal route to the goal. We could supervise the adapter with that distance, but doing so would build our knowledge of these particular environments into the method. For other tasks, the right measure of progress may be difficult to define---for example, when a robot must arrange scattered blocks into a target shape. We therefore propose a task-agnostic training procedure that aims to discover a useful measure of progress from experience, rather than fitting an adapter to targets specified using our expert knowledge of each task.

Our approach builds on an observation: close to the goal, the raw latent cost often has enough structure to guide the planner successfully. We first collect trajectories from starting positions near the goal. A successful trajectory eventually reaches the goal, so the order of its observations provides a weak signal of progress: observations visited later tend to be closer to completing the task than those visited earlier. We fit the adapter so that later observations have lower adapted latent cost to the goal, while recognizing that this ordering need not hold at every step. We then collect trajectories from progressively farther starting positions and repeat, using each updated adapter to guide the next stage.

In matched 50-case planning evaluations, the adapted latent cost increases success from 54\% to 98\% in DotWall and from 82\% to 96\% in PointMaze. We also evaluate the DotWall adapter on 16 cases from unseen environment layouts, where the wall and doorway have moved from their positions during training. Its latent cost agrees more closely with geodesic distance on these layouts, improving legal-path pair ordering from 71.9\% to 85.8\%.

Our contributions are:
\begin{enumerate}
    \item We show how raw I-JEPA latent cost can misguide a CEM-based planner and measure the resulting planning failures.
    \item We propose a task-agnostic curriculum for fitting a small, task-specific latent adapter from collected trajectories, without supervision on expert targets.
    \item We show that the adapter improves planning in two environments. Moreover, the adapted latent cost in DotWall generalizes to unseen environment layouts: it agrees closely with the geodesic distance that describes progress in this task, even when the wall and doorway change position.
\end{enumerate}

\section{Related Work}
Early visual world models learned representations and dynamics to support control. World Models trained a visual autoencoder and a recurrent dynamics model, then optimized a small controller for the task reward \cite{ha2018worldmodels}. PlaNET learned a latent dynamics model together with observation and reward models, and used the cross-entropy method (CEM) to plan online \cite{hafner2019planet}. Dreamer took a different approach: it learned an actor and a value function from trajectories imagined by its world model, rather than searching over action sequences at every decision \cite{hafner2020dreamer}. DreamerV2 extended this approach with discrete latent states, while DreamerV3 demonstrated its effectiveness across a much wider range of tasks \cite{hafner2021dreamerv2,hafner2025dreamerv3}. TD-MPC is closer to online planning: it combines short-horizon planning in a learned latent space with a value estimate for outcomes beyond that horizon \cite{hansen2022tdmpc}. These methods show that latent world models can support effective control. Their reward predictors, value functions, or policies are nevertheless tied to the task objective. Changing the goal does not necessarily require relearning the environment dynamics, but it may require adapting those goal-dependent components or the planning objective.

Image-goal planning offers another way to specify what the agent should reach. DINO-WM and V-JEPA 2-AC predict future visual representations and compare them with the representation of a goal image when evaluating actions \cite{zhou2025dinowm,assran2025vjepa2}. This makes it possible to change the goal image without training a new reward predictor for each goal. It also makes the distance in visual latent space an important part of the planner: a low latent cost is only helpful if it indicates progress that the agent can actually make.

Other goal-reaching methods have learned such information from experience. Search on the Replay Buffer (SoRB) learns a goal-conditioned measure of reachability and uses it to find routes through previously visited observations \cite{eysenbach2019sorb}. Contrastive RL learns representations whose similarity corresponds to a goal-conditioned value function \cite{eysenbach2022contrastive}. These works use different policies and planning procedures from ours, but they show why visual similarity alone need not be the right measure of distance to a goal. \cite{alrasheed2026planninglimitslatentworld} explores the failure of latent planning in more detail.

Temporal Straightening addresses the latent planning cost more directly by training an encoder and predictor so that feasible trajectories become straighter in latent space \cite{wang2026straightening}. We address the same difficulty while keeping the pretrained I-JEPA encoder and the learned world model fixed. Instead of training a new encoder or supervising an adapter with shortest-path distances, we fit a small adapter using comparisons obtained from collected trajectories. In DotWall and PointMaze, geodesic distance lets us evaluate the resulting latent cost, but it is not supplied as a training target.

\begin{figure}[t]
\centering
\includegraphics[width=\columnwidth]{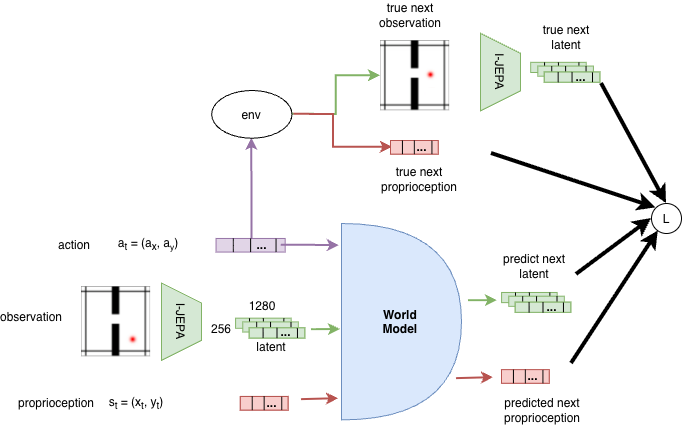}
\caption{World-model training. Frozen I-JEPA encodes the current and next images. The action-conditioned world model predicts the next visual embeddings and proprioceptive state, which are compared with the observed targets.}
\label{fig:training-compact}
\end{figure}

\begin{figure*}[t]
\centering
\includegraphics[width=\textwidth]{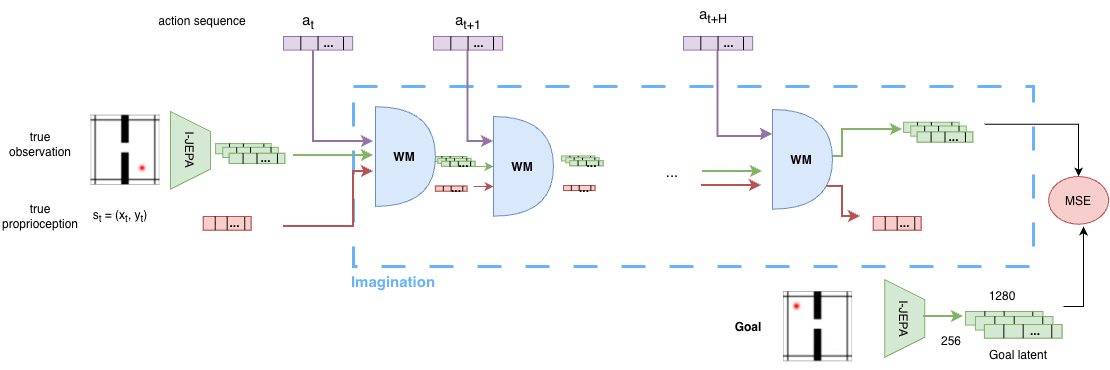}
\caption{Image-goal planning with the learned world model. CEM proposes action sequences, the model predicts their outcomes in latent space, and the MSE between a predicted terminal latent and the encoded goal scores each sequence. Only the first selected action is executed before replanning.}
\label{fig:planning-latent}
\end{figure*}

\section{World Model and Latent Planning}
\subsection{World-model training}
Our world model follows a common pattern in recent visual world modeling: a visual encoder produces a latent representation of an observation, and an action-conditioned predictor estimates how that representation changes. Related architectures are used in DINO-WM and V-JEPA 2-AC, although their encoders and details might differ. \cite{zhou2025dinowm,assran2025vjepa2}. Details about the internal components of our world model can be found in \ref{app:wm-training}

Let $o_t$ be an image observation, $s_t$ the ball's proprioceptive state, and $a_t$ an action. A frozen I-JEPA ViT-H/14 encoder $E$ splits each $224\times224$ image into $16\times16=256$ patches and produces one 1280-dimensional latent embedding per patch. The transformer-based world model $F$ receives these embeddings $z_t=E(o_t)$, $s_t$, and $a_t$, then predicts the next latent and proprioceptive state, $(\hat z_{t+1},\hat s_{t+1})=F(z_t,s_t,a_t)$. To train it, we encode the observed next image with the same frozen encoder and compare the predicted embeddings with those targets. We also compare its predicted proprioceptive state with the observed next state. Fig.~\ref{fig:training-compact} summarizes this training setup. Details about the construction of the datasets used for training the world model can be found in \ref{app:datasets}.

\subsection{Planning toward an image goal}
Once trained, the world model predicts the outcome of actions in latent space. The current observation and goal image $o_g$ are encoded by I-JEPA, with $z_g=E(o_g)$. At each decision, the cross-entropy method (CEM) searches over candidate action sequences. The world model rolls out each sequence in latent space, and the predicted final embeddings are scored against $z_g$. The planner executes the first action of the selected sequence, observes the new state, and plans again. This is closed-loop model-predictive control.

The raw latent cost is the mean squared error (MSE) between corresponding spatial embeddings and their channels:
\begin{equation}
c_{\rm raw}(z,z_g)=\frac{1}{ND}\sum_{i=1}^{N}\sum_{j=1}^{D}(z_{ij}-z_{g,ij})^2,
\label{eq:raw}
\end{equation}
where $N=256$ and $D=1280$ for I-JEPA. Thus every patch embedding contributes to one scalar latent cost. Fig.~\ref{fig:planning-latent} shows how the world model is rolled out repeatedly in imagination, while the goal image enters through the terminal latent cost used to select a plan. The encoder and world model remain fixed throughout our experiments.

\subsection{When raw latent cost misleads planning}
We study two image-goal navigation environments, shown in Fig.~\ref{fig:environments}. \emph{DotWall} has a ball, a separating wall, and a doorway; its standard layout has wall coordinate $x=32$ and doorway coordinate $y=30$. \emph{PointMaze} is a top-down U-shaped maze. In both tasks the ball must reach the goal position, but the legal route can initially move away from the goal in Euclidean distance or visual similarity.

\begin{figure}[t]
\centering
\setlength{\tabcolsep}{1pt}
\begin{tabular}{@{}ccc@{}}
{\scriptsize DINOv2} & {\scriptsize I-JEPA} & {\scriptsize Legal path}\\
\multicolumn{3}{c}{\scriptsize DotWall, goal $(15,15)$}\\[-1mm]
\includegraphics[width=.29\columnwidth]{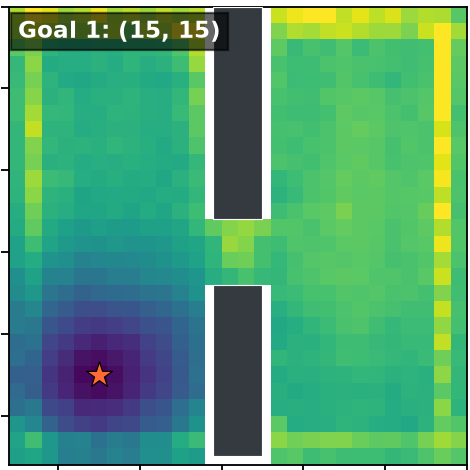} &
\includegraphics[width=.29\columnwidth]{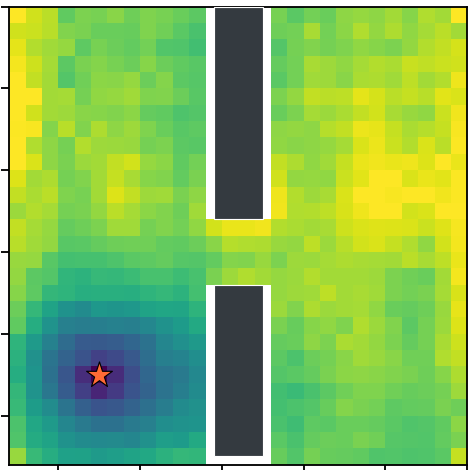} &
\includegraphics[width=.29\columnwidth]{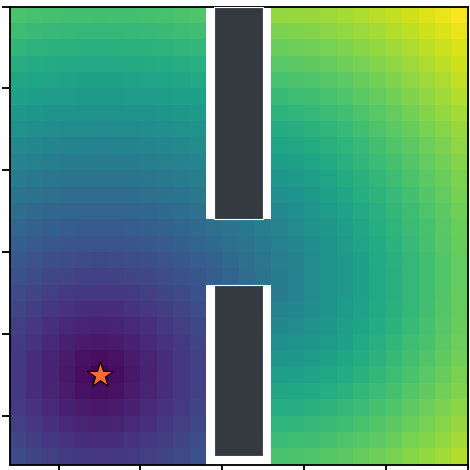}\\[1mm]
\multicolumn{3}{c}{\scriptsize PointMaze, goal $(-1.09,0.81)$}\\[-1mm]
\includegraphics[width=.29\columnwidth]{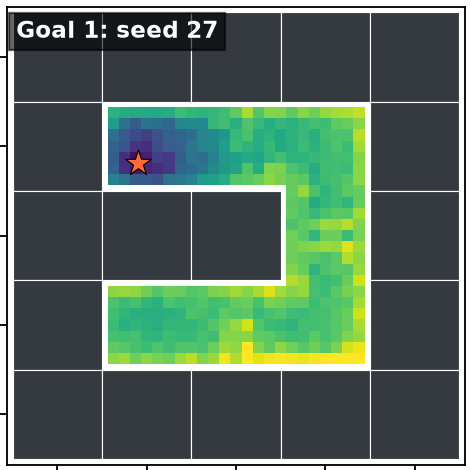} &
\includegraphics[width=.29\columnwidth]{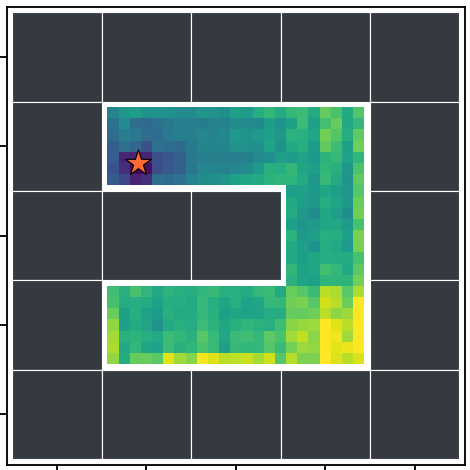} &
\includegraphics[width=.29\columnwidth]{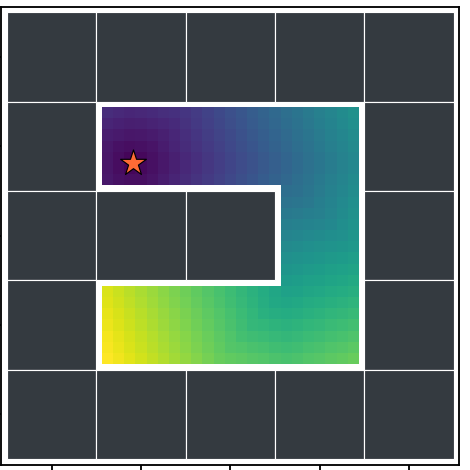}
\end{tabular}
\caption{Raw latent-cost heatmaps for one goal in each environment, compared with shortest legal-path distance. Darker colors indicate lower values; each panel has its own color scale. The costs are computed from rendered observations, not imagined world-model rollouts. Both encoders show departures from legal-path ordering, although their relative quality varies by goal.}
\label{fig:raw-costs}
\end{figure}

The planning procedure depends on the latent cost giving useful guidance. To inspect that cost independently of world-model prediction errors, we render an observation with the ball at each legal position and compare its encoded representation with that of a fixed goal image. Each cell in Fig.~\ref{fig:raw-costs} shows the MSE between the latent representations of these two images; the star marks the goal. To examine whether the problem extends beyond one pretrained encoder, we compute the heatmaps with both I-JEPA ViT-H/14 and DINOv2 ViT-S/14. Their latent-cost scales differ, so we compare the ordering of positions within each map rather than color magnitudes across encoders.

The maps show that neither pretrained encoder consistently orders states according to the travel required to reach the goal. The raw latent cost is often informative close to the goal, but farther away it can favor a visually similar position across a wall or fail to reward movement toward the only legal passage. The two encoders differ in how strongly they exhibit this problem for individual goals; raw visual similarity does not reliably encode progress along a legal route.

This creates a difficulty for a search-based planner with a finite horizon. To reach the goal, an action sequence may first have to pass through a doorway or around an obstacle, often temporarily increasing the raw latent cost. If the planning horizon ends before the sequence reaches a state with lower cost, its necessary first actions may look worse than actions that make no useful progress. A longer horizon can expose the benefit of the detour, but it also enlarges the action-sequence search space and compounds world-model prediction errors. We therefore seek a latent cost that provides useful guidance over more of the state space, without requiring the planner to discover every detour through a long imagined rollout.

\section{Method}
\subsection{Fitting an adapter for latent goal scoring}
\begin{figure}[t]
\centering
\includegraphics[width=.96\columnwidth]{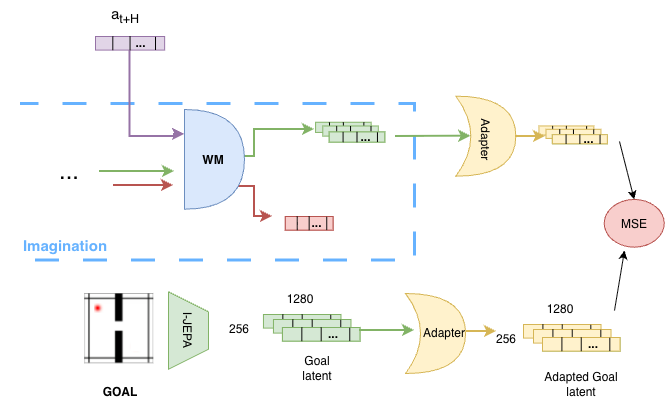}
\caption{The encoder and world model remain frozen. The adapter transforms predicted and goal embeddings only on the latent goal-cost path used by CEM.}
\label{fig:architecture}
\end{figure}

The useful visual features produced by the pretrained I-JEPA encoder enable our world model to accurately predict dynamics in latent space. The difficulty lies in how the planner scores those predictions. We therefore keep the encoder and world model frozen and fit a small adapter that transforms their latent representations before the planner compares a predicted terminal latent with the goal (Fig.~\ref{fig:architecture}). The adapter is applied to both representations, and their mean-squared error (MSE) becomes the adapted latent cost used to score candidate action sequences.

In DotWall and PointMaze, an obvious solution would be to calculate geodesic distance to the goal and train the adapter to reproduce it. This relies on knowledge specific to these navigation tasks. For other tasks, an appropriate measure of progress may be difficult to specify by hand. We therefore neither fit the adapter on geodesic targets nor use geodesic improvement to select training examples. Instead, we aim to adapt the latent cost from experience without expert guidance about which individual states constitute progress. Geodesic distance serves only as an evaluation reference in these two environments.

We apply a shared residual multilayer perceptron $A_\theta$ independently to each spatial embedding. It has a 64-dimensional bottleneck and is initialized as the identity map. Its output replaces the raw embeddings only in the goal-cost computation:
\begin{equation}
c_\theta(z,z_g)=\frac{1}{ND}\sum_{i=1}^{N}\sum_{j=1}^{D}
\big(A_\theta(z_i)_j-A_\theta(z_{g,i})_j\big)^2.
\label{eq:adapted}
\end{equation}
The world model continues to receive unmodified encoder embeddings, and both $E$ and $F$ remain frozen. Identity initialization makes the initial planner equivalent to raw latent-MSE planning. The same adapter is applied to predicted candidate embeddings and the goal embeddings at evaluation time.

\subsection{Collection phase}
We observe that raw I-JEPA latent cost often provides a useful planning signal close to the goal and we want to take advantage of this property and gradually diffuse it further away. We begin with planning tasks whose starting positions lie in a goal-centered square region, which we call Band 1. Initially, the agent plans with raw latent MSE using the original planning architecture (Fig.~\ref{fig:planning-latent}). During execution, we record the visited observations and whether the agent reaches the goal.

Successful trajectories provide the main training signal. An agent that reaches the goal has, over the course of that trajectory, made progress toward it. We therefore use the order of visited observations to form comparisons: an observation visited later should generally receive a lower adapted latent cost to the goal than one visited earlier on the same route. This is an imperfect assumption, because a successful route can include detours or loops. However in practice most successful trajectories went directly to the goal and our method performed well despite the noisy temporal comparisons introduced by the detours.


Failed trajectories can also provide useful training signal because some make partial progress. We rank sampled observations from a failed episode by their latent-cost proximity to successful routes for the same goal, and form additional pairs from that ordering. This approach is explained in more detail in the appendix \ref{app:failed-episodes}.

\subsection{Fitting phase and expanding bands}
We fit the adapter on pairs collected from successful and failed trajectories. Let $e$ and $l$ be observations that a pair ranks as worse and better, respectively, for goal $g$. We penalize the adapter when $l$ does not have a lower adapted latent cost than $e$ by a margin $m$:

\begin{equation}
\mathcal L_{\rm rank}(e,l,g)=
\max\{0,\,m+d_{\theta}(e, l, g)\}.
\label{eq:rank}
\end{equation}
where:
\begin{equation*}
 d_{\theta}(e,l,g)=
c_\theta(E(l),E(g))-c_\theta(E(e),E(g))
\end{equation*}
For successful trajectories, $e$ and $l$ were visited earlier and later; for failed trajectories, their order comes from successful-route matching.

\begin{figure}[t]
\centering
\includegraphics[width=.96\columnwidth]{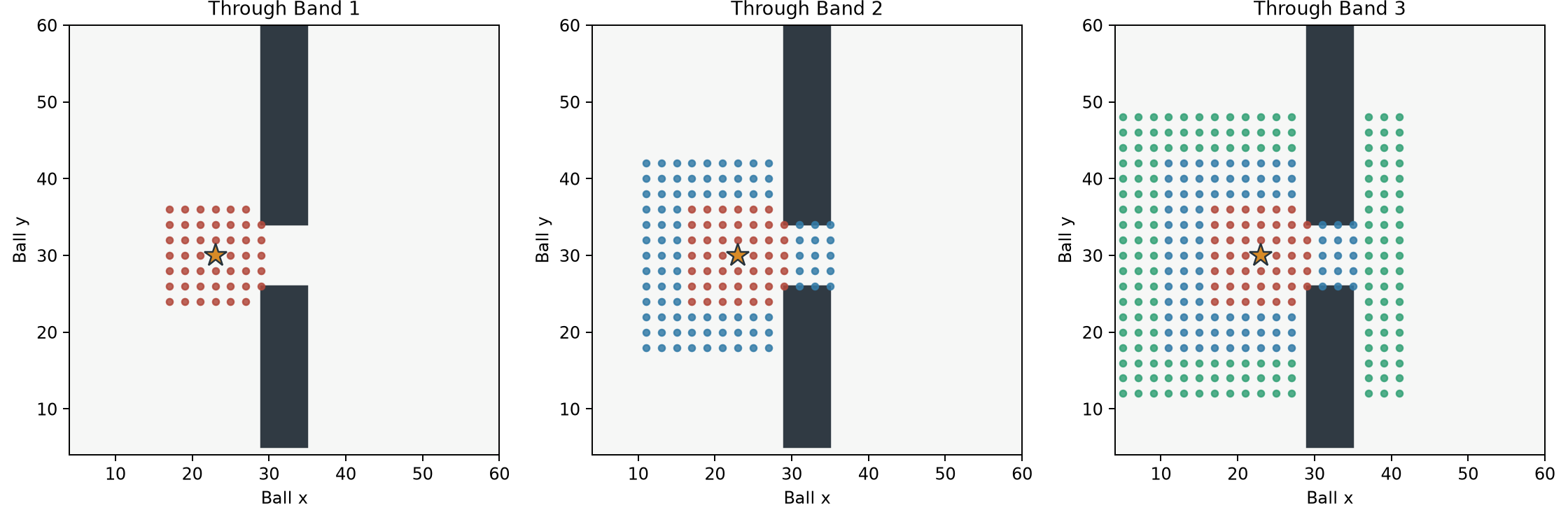}
\caption{The first three DotWall start bands around one goal. Each stage adds starts farther from the goal. Bands determine starting positions, not the region an episode must remain within.}
\label{fig:bands}
\end{figure}

Within each band, we alternate collection and fitting. The adapter remains frozen during collection. After each round, we add new pairs to cumulative replay and fit the adapter; its updated latent cost guides the next round. Once Band 1 is complete, we add starting positions farther from the goal in Band 2 and continue expanding (Fig.~\ref{fig:bands}) until the chosen start grid covers the state space.

\subsection{Filtering collected experience}
Temporal order does not always describe progress: an agent may leave a region and return to it before eventually succeeding. We remove the intervening loop when a trajectory returns near an earlier visited position. We then form successful-route comparisons only from consecutive recorded observations whose endpoints remain after filtering. We also discard unchanged observations. The bands constrain starting positions; observations need not remain inside the current band.
These rules clean the trajectory data without checking whether a step reduces shortest legal-path distance. The method uses a goal-success signal and a way to generate starting states around a goal, but it does not use an oracle progress measure to label the collected pairs.

\section{Experiments}
We evaluate the adapted latent cost in DotWall and PointMaze. In both environments, we measure closed-loop planning success  and compare it with raw I-JEPA latent based planning on matched start--goal cases, then inspect whether the latent cost orders legal positions in a way that helps the agent reach the goal.

\subsection{Evaluation protocol}
\begin{figure}[t]
\centering
\includegraphics[width=.75\columnwidth]{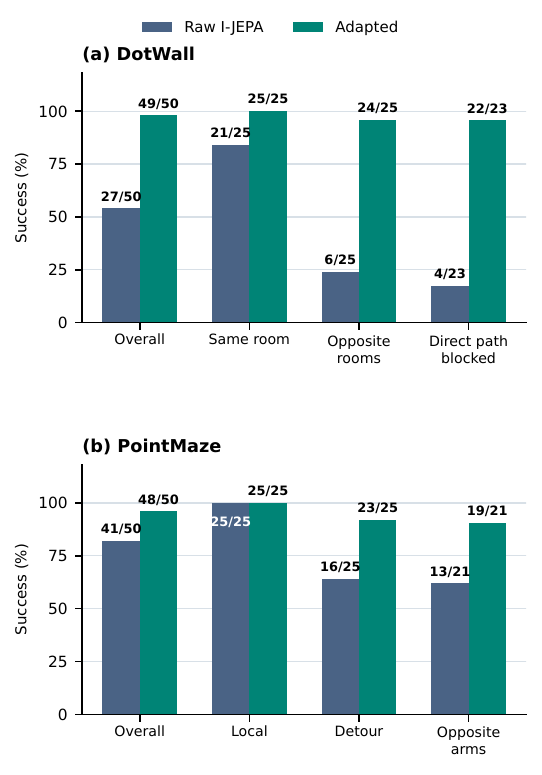}
\caption{Matched closed-loop planning success with raw and adapted I-JEPA latent costs. (a) DotWall; directly blocked cases are a subset of opposite-room cases. (b) PointMaze; opposite-arm cases are a subset of detours. Counts above bars are successes over evaluated cases.}
\label{fig:planning-success}
\end{figure}

Each raw-versus-adapted planning comparison uses the same frozen I-JEPA encoder, world model, CEM settings, start, goal, and planner seed, with a limit of 100 simulator steps. The 50 cases in each environment use 10 goals held out from adapter fitting. Figure~\ref{fig:planning-success} gives the paired success rates; the fractions above its bars give the number of successful cases. The subset categories overlap, so their counts should not be added to the overall total.

\begin{figure}[t]
\centering
\includegraphics[width=\columnwidth]{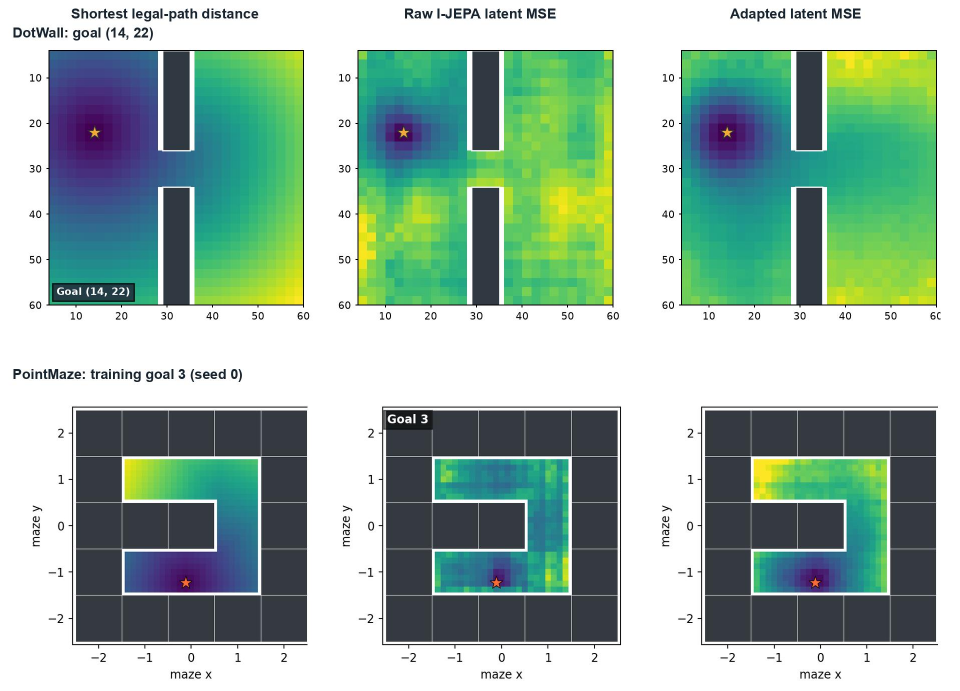}
\caption{Latent-cost fields for one goal in each environment. Columns show shortest legal-path distance, raw I-JEPA latent MSE, and adapted latent MSE. Each position represents the cost of a rendered observation with the ball at that position relative to the goal image (star). DotWall goal $(14,22)$ was held out from adapter fitting; Each panel has an independent color scale.}
\label{fig:experiment-costs}
\end{figure}

To inspect the latent cost independently of world-model prediction error, we render an observation at each legal ball position and compare it with a fixed goal image. Figure~\ref{fig:experiment-costs} shows one goal per environment. The shortest legal-path distance is also shown for comparison. The columns use different numerical scales; their structure is what we are interested in rather than the exact magnitudes.

\begin{figure}[b]
\centering
\includegraphics[width=.94\columnwidth]{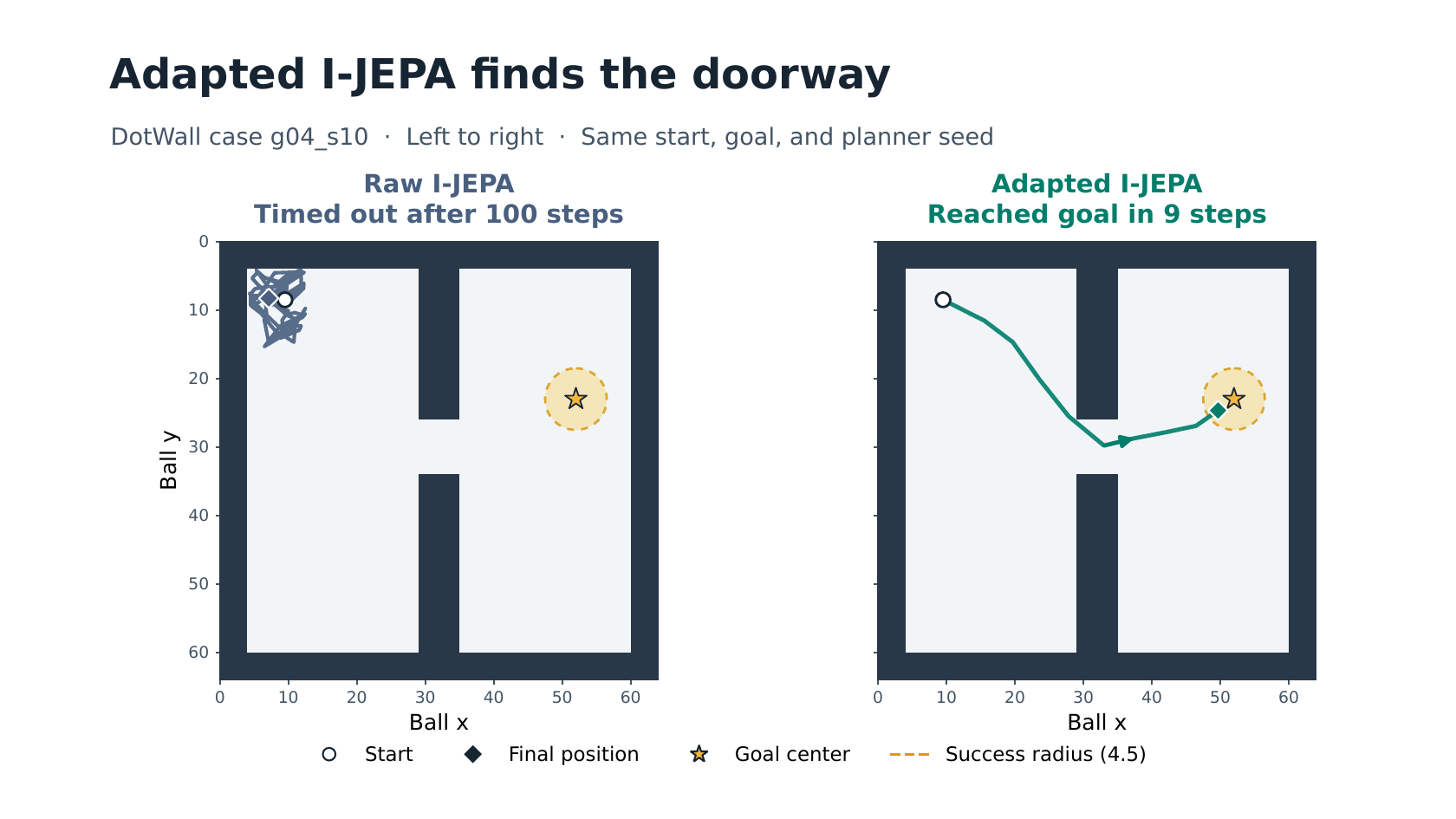}
\caption{Matched DotWall example: raw-cost planning times out in the left room; adapted-cost planning crosses the doorway and reaches the goal.}
\label{fig:dotwall-example}
\end{figure}

\subsection{DotWall}
The adapter was fitted on the fixed DotWall layout with the wall at $x=32$ and doorway centered at $y=30$. Collection used 12 training goals in both rooms. Legal starting positions were drawn from a two-unit lattice in eight goal-centered square bands; their half-widths increased from 6 to 48 units in steps of 6 (Fig.~\ref{fig:bands}). We used three collection--fitting rounds per band. Across all rounds, 6,121 rollouts produced 5,721 successful episodes. In the outermost band, collection succeeded from 104 of 116 starts, including 82 of 94 starts opposite the goal room. These are collection outcomes under successive adapter checkpoints, distinct from the matched test below.

On the 50 held-out planning cases, success rises from 27/50 (54\%) with raw I-JEPA latent MSE to 49/50 (98\%) with the adapted cost (Fig.~\ref{fig:planning-success}a). Same-room success rises from 21/25 to 25/25, while opposite-room success rises from 6/25 to 24/25. When the straight path to the goal is blocked by the wall, success rises from 4/23 to 22/23. All 27 raw successes remain successes after adaptation; the adapter solves 22 additional cases.

The first row of Fig.~\ref{fig:experiment-costs} explains the difficulty behind the opposite-room results. Raw I-JEPA MSE is low around the goal, but its field away from the goal does not consistently reward the route through the doorway. The adapted cost has a smoother decrease along legal approaches. In the matched example of Fig.~\ref{fig:dotwall-example}, raw-cost planning remains in the starting room until timeout, whereas adapted-cost planning crosses the doorway and reaches the goal using the same world model and CEM settings.

\subsection{PointMaze}
We use the same collection and fitting rule in a different environment, the U-shaped PointMaze. Training used four goals and 632 collection episodes from legal starts across seven bands, with three collection--fitting rounds per band. The final adapter is the checkpoint after Band 7, Round 3; the encoder and PointMaze world model remain frozen.

On 50 matched cases using 10 goals held out from fitting, raw-cost planning succeeds in 41/50 (82\%) and adapted-cost planning in 48/50 (96\%) (Fig.~\ref{fig:planning-success}b). Both costs solve all 25 local cases. The gain is in detours around the central wall: success increases from 16/25 to 23/25, including an increase from 13/21 to 19/21 for the opposite-arm subset. 

\begin{figure}[t]
\centering
\includegraphics[width=.94\columnwidth]{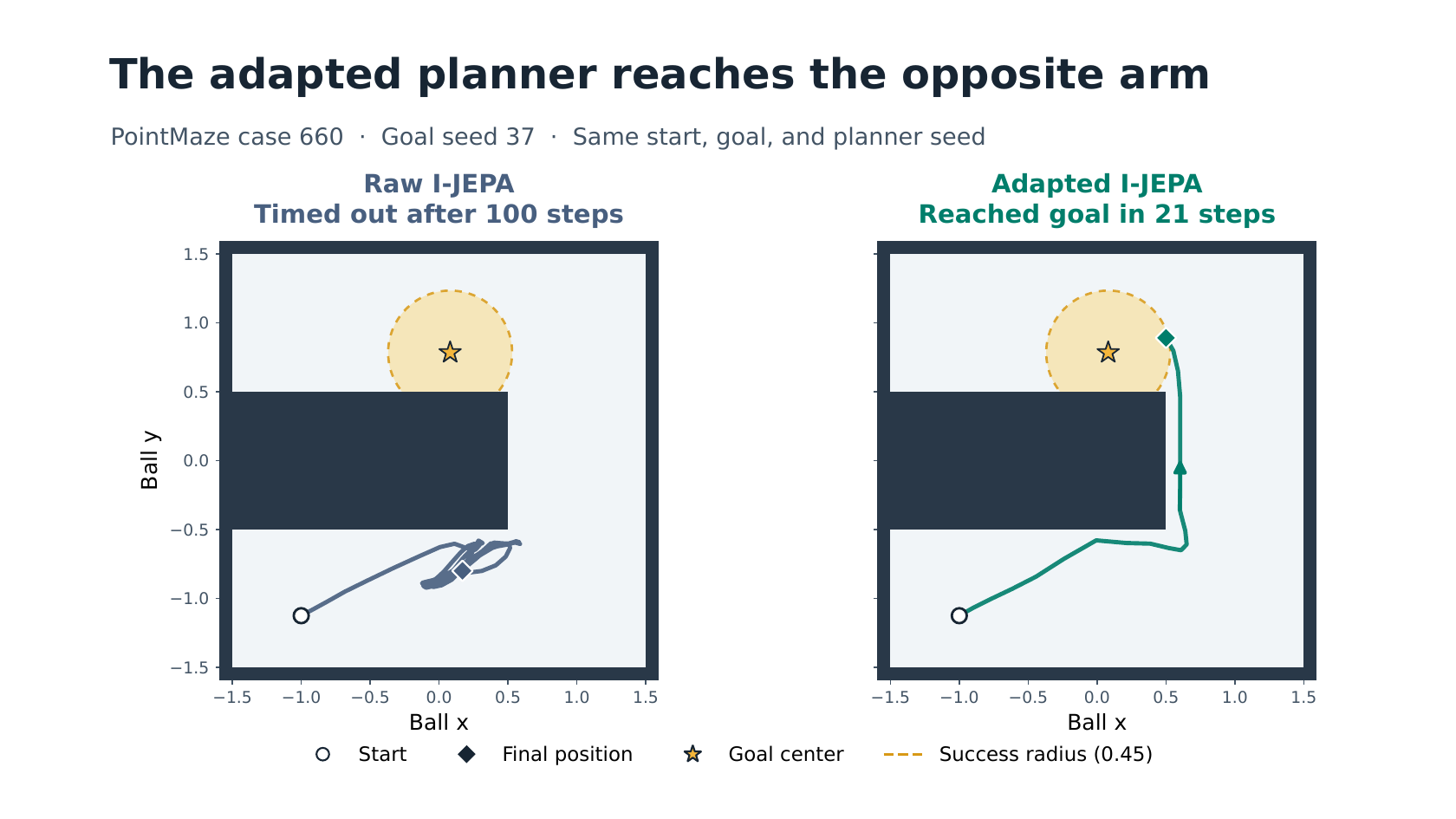}
\caption{Matched PointMaze detour: raw-cost planning remains in the lower arm until timeout, while adapted-cost planning goes around the central wall and reaches the goal.}
\label{fig:pointmaze-example}
\end{figure}

In the second row of Fig.~\ref{fig:experiment-costs}, the raw latent MSE is irregular along the corridor and weakly distinguishes positions that require a long detour from those near the goal. After adaptation, cost increases more consistently away from the goal along the legal route. Figure~\ref{fig:pointmaze-example} shows one held-out detour: raw-cost planning remains in the lower arm until timeout; adapted-cost planning travels around the obstacle and reaches the upper-arm goal.

\subsection{Generalization to unseen DotWall layouts}
We next test whether the DotWall adapter trained with a fixed wall and doorway gives useful latent costs when those obstacles move. We freeze the adapter fitted only on the standard layout $(x_{\rm wall},y_{\rm door})=(32,30)$ and evaluate four unseen layouts: $(26,22)$, $(26,38)$, $(38,22)$, and $(38,38)$. Four goals held out from adapter fitting are tested on each layout, yielding 16 layout--goal cases. For each case, we render 688 legal ball positions and compare raw and adapted latent costs with shortest legal-path distance.

Table~\ref{tab:transfer} reports four aspects of this agreement. Map-wide pair ordering is the fraction of position pairs for which the position with shorter legal-path distance has lower latent cost, restricting pairs to a distance difference of at least 0.5 units. Opposite-room ordering applies the same test to positions across the wall from the goal. Spearman $\rho$ compares the rankings of latent cost and legal-path distance across all positions. Doorway-local uphill steps are steps near the doorway that shorten the legal path while increasing latent cost, a failure mode for a greedy cost signal. Higher values are better for the first three rows, while lower is better for the last. The first three rows average case-level scores; the final row pools doorway-local steps across cases.

\begin{table}[t]
\centering
\caption{Latent-cost geometry on 16 unseen DotWall layout--goal cases.}
\label{tab:transfer}
\begin{tabular}{lrr}
\toprule
Measure & Raw I-JEPA & Adapted \\
\midrule
Map-wide pair ordering & 71.9\% & \textbf{85.8\%} \\
Opposite-room pair ordering & 48.5\% & \textbf{77.1\%} \\
Mean Spearman $\rho$ & 0.594 & \textbf{0.864} \\
Doorway-local uphill steps & 39.2\% & \textbf{15.6\%} \\
\bottomrule
\end{tabular}
\end{table}

\begin{figure}[t]
\centering
\includegraphics[width=.75\columnwidth]{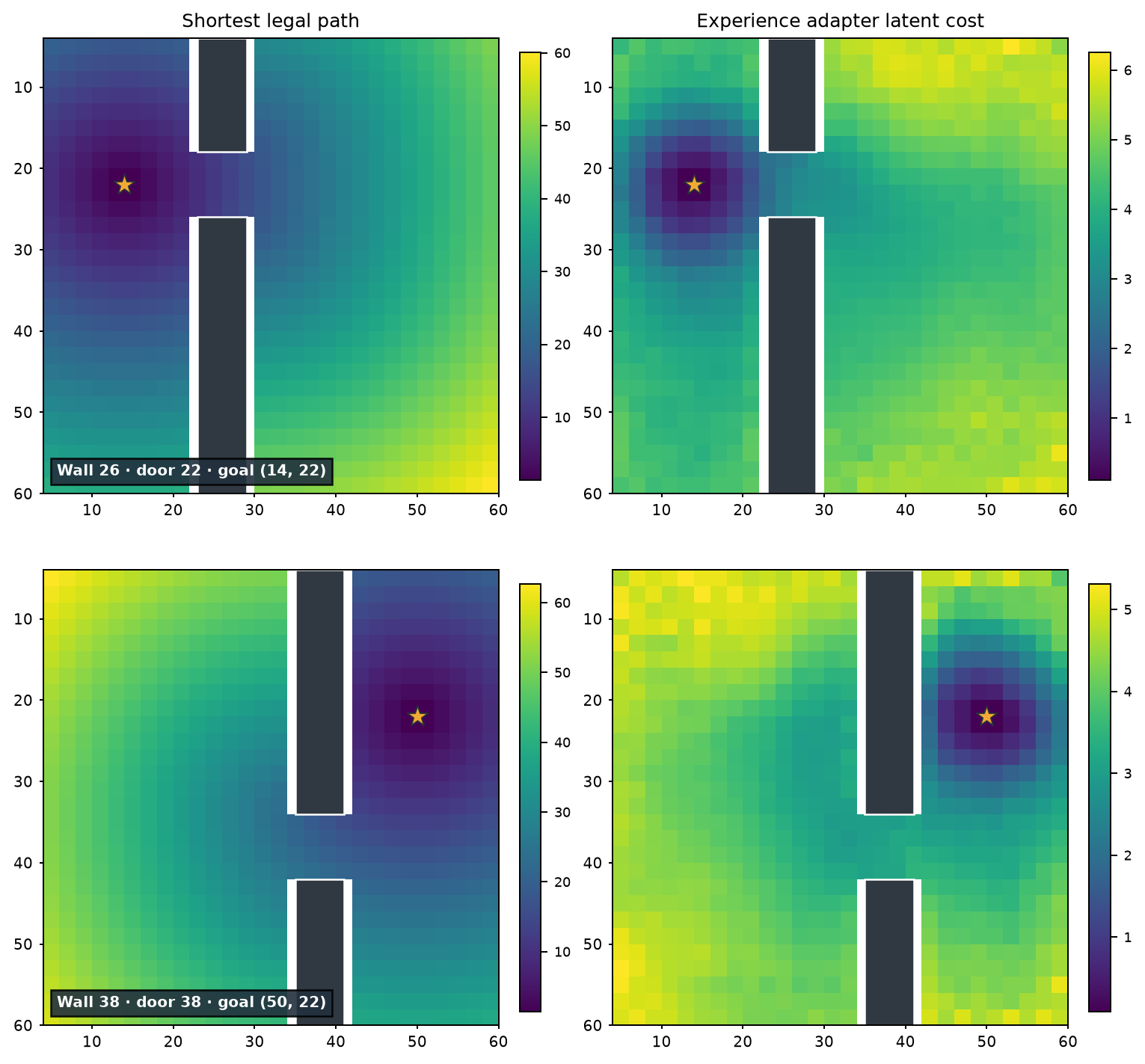}
\caption{Two unseen DotWall layouts (rows), with shortest legal-path distance and adapted latent cost (columns). The adapter was fitted only on the standard $(32,30)$ layout. Each panel uses its own color scale.}
\label{fig:transfer}
\end{figure}

The adapted cost improves map-wide pair ordering on all 16 unseen cases. Its mean ordering rises from 71.9\% to 85.8\%, while opposite-room ordering rises from 48.5\% to 77.1\% (Table~\ref{tab:transfer}). Figure~\ref{fig:transfer} shows two unseen layouts: the adapted fields retain a route-shaped ordering that broadly follows the legal-path reference despite the changed wall and doorway. This is a transfer test of the latent-cost field. 

\section{Discussion and Limitations}
The paired planning gains point to a concrete weakness of raw visual similarity as a goal objective in obstacle-rich settings. Fitting an adapter from gradually broader planning experience can improve the objective while retaining the pretrained encoder and world model. The unseen-layout diagnostic suggests that the learned adapter transfers its benefits to unseen layouts when the obstacle geometry changes. 

The method still requires a goal-success signal and enough coverage in each band. Later bands require more starting positions to cover each region. In state spaces with more dimensions than the two ball coordinates, the starting positions required increase rapidly with respect to the number of dimensions. This curse of dimensionality may make the curriculum expensive or sparse. Temporal rankings are only weak labels: even successful routes can detour, and failure-route matching used to rank observations in failed trajectories relies on the current adapter.


\section{Conclusion}
Raw pretrained visual embeddings can produce a locally useful but globally misleading image-goal cost. We showed that a small adapter, fitted through expanding collection and temporal trajectory rankings, improves matched planning success in DotWall and PointMaze while keeping the encoder and world model frozen. The adapter also improves latent-cost ordering on unseen DotWall layouts, transferring its benefits when the map's geometry changes. Our method uses no hand-crafted supervision targets such as geodesic distance which would tie it to the specific navigation tasks. Instead it uses success signal and collected experience which is cleaned up with minimal filtering. We leave the deployment of our method in higher dimensionality tasks for future work.

\bibliographystyle{IEEEtran}
\bibliography{references}

@inproceedings{assran2023ijepa,
  author = {Mahmoud Assran and Quentin Duval and Ishan Misra and Piotr Bojanowski and Pascal Vincent and Michael Rabbat and Yann LeCun and Nicolas Ballas},
  title = {Self-Supervised Learning from Images with a Joint-Embedding Predictive Architecture},
  booktitle = {Proceedings of the IEEE/CVF Conference on Computer Vision and Pattern Recognition},
  pages = {15619--15629},
  year = {2023}
}

@article{oquab2024dinov2,
  author = {Maxime Oquab and Timoth{\'e}e Darcet and Th{\'e}o Moutakanni and Huy V. Vo and Marc Szafraniec and Vasil Khalidov and Pierre Fernandez and Daniel Haziza and Francisco Massa and Alaaeldin El-Nouby and Mido Assran and Nicolas Ballas and Wojciech Galuba and Russell Howes and Po-Yao Huang and Shang-Wen Li and Ishan Misra and Michael Rabbat and Vasu Sharma and Gabriel Synnaeve and Hu Xu and Herv{\'e} J{\'e}gou and Julien Mairal and Patrick Labatut and Armand Joulin and Piotr Bojanowski},
  title = {{DINOv2}: Learning Robust Visual Features without Supervision},
  journal = {Transactions on Machine Learning Research},
  year = {2024}
}

@inproceedings{zhou2025dinowm,
  author = {Gaoyue Zhou and Hengkai Pan and Yann LeCun and Lerrel Pinto},
  title = {{DINO-WM}: World Models on Pre-trained Visual Features Enable Zero-shot Planning},
  booktitle = {Proceedings of the 42nd International Conference on Machine Learning},
  volume = {267},
  series = {Proceedings of Machine Learning Research},
  pages = {79115--79135},
  year = {2025}
}

@inproceedings{hafner2019planet,
  author = {Danijar Hafner and Timothy Lillicrap and Ian Fischer and Ruben Villegas and David Ha and Honglak Lee and James Davidson},
  title = {Learning Latent Dynamics for Planning from Pixels},
  booktitle = {Proceedings of the 36th International Conference on Machine Learning},
  volume = {97},
  series = {Proceedings of Machine Learning Research},
  pages = {2555--2565},
  year = {2019}
}

@inproceedings{hafner2020dreamer,
  author = {Danijar Hafner and Timothy Lillicrap and Jimmy Ba and Mohammad Norouzi},
  title = {Dream to Control: Learning Behaviors by Latent Imagination},
  booktitle = {International Conference on Learning Representations},
  year = {2020}
}

@inproceedings{ha2018worldmodels,
  author = {David Ha and J{\"u}rgen Schmidhuber},
  title = {Recurrent World Models Facilitate Policy Evolution},
  booktitle = {Advances in Neural Information Processing Systems},
  volume = {31},
  year = {2018},
  url = {https://proceedings.neurips.cc/paper/2018/hash/2de5d16682c3c35007e4e92982f1a2ba-Abstract.html}
}

@misc{assran2025vjepa2,
  author = {Mahmoud Assran and Adrien Bardes and David Fan and others},
  title = {{V-JEPA 2}: Self-Supervised Video Models Enable Understanding, Prediction and Planning},
  year = {2025},
  eprint = {2506.09985},
  archivePrefix = {arXiv},
  url = {https://arxiv.org/abs/2506.09985}
}

@inproceedings{wang2026straightening,
  author = {Ying Wang and Oumayma Bounou and Gaoyue Zhou and Randall Balestriero and Tim G. J. Rudner and Yann LeCun and Mengye Ren},
  title = {Temporal Straightening for Latent Planning},
  booktitle = {Proceedings of the 43rd International Conference on Machine Learning},
  volume = {306},
  series = {Proceedings of Machine Learning Research},
  pages = {125168--125196},
  year = {2026},
  url = {https://proceedings.mlr.press/v306/wang26n.html}
}

@inproceedings{hafner2021dreamerv2,
  author = {Danijar Hafner and Timothy Lillicrap and Mohammad Norouzi and Jimmy Ba},
  title = {Mastering Atari with Discrete World Models},
  booktitle = {International Conference on Learning Representations},
  year = {2021},
  url = {https://openreview.net/forum?id=0oabwyZbOu}
}

@article{hafner2025dreamerv3,
  author = {Danijar Hafner and Jurgis Pasukonis and Jimmy Ba and Timothy Lillicrap},
  title = {Mastering Diverse Control Tasks Through World Models},
  journal = {Nature},
  volume = {640},
  pages = {647--653},
  year = {2025},
  doi = {10.1038/s41586-025-08744-2}
}

@inproceedings{hansen2022tdmpc,
  author = {Nicklas A. Hansen and Hao Su and Xiaolong Wang},
  title = {Temporal Difference Learning for Model Predictive Control},
  booktitle = {Proceedings of the 39th International Conference on Machine Learning},
  series = {Proceedings of Machine Learning Research},
  volume = {162},
  pages = {8387--8406},
  year = {2022},
  url = {https://proceedings.mlr.press/v162/hansen22a.html}
}

@inproceedings{eysenbach2019sorb,
  author = {Ben Eysenbach and Ruslan Salakhutdinov and Sergey Levine},
  title = {Search on the Replay Buffer: Bridging Planning and Reinforcement Learning},
  booktitle = {Advances in Neural Information Processing Systems},
  volume = {32},
  year = {2019},
  url = {https://proceedings.neurips.cc/paper/2019/hash/5c48ff18e0a47baaf81d8b8ea51eec92-Abstract.html}
}

@inproceedings{eysenbach2022contrastive,
  author = {Benjamin Eysenbach and Tianjun Zhang and Sergey Levine and Ruslan Salakhutdinov},
  title = {Contrastive Learning as Goal-Conditioned Reinforcement Learning},
  booktitle = {Advances in Neural Information Processing Systems},
  volume = {35},
  year = {2022},
  url = {https://proceedings.neurips.cc/paper_files/paper/2022/hash/e7663e974c4ee7a2b475a4775201ce1f-Abstract-Conference.html}
}

@misc{alrasheed2026planninglimitslatentworld,
      title={The Planning Limits of Latent World Models}, 
      author={Ali Alrasheed and Basim Azam and Naveed Akhtar},
      year={2026},
      eprint={2609.39235},
      archivePrefix={arXiv},
      primaryClass={cs.RO},
      url={https://arxiv.org/abs/2609.39235}, 
}

\clearpage
\onecolumn

\noindent{\LARGE\bfseries A. Appendix}\par
\vspace{1em}

\newcounter{appsub}
\renewcommand{\theappsub}{A.\arabic{appsub}}
\newcommand{\appsubsection}[1]{%
  \refstepcounter{appsub}%
  \par\bigskip\noindent{\Large\bfseries \theappsub. #1}\par\medskip
}

\appsubsection{World-model architecture and training}
\label{app:wm-training}


\begin{figure}[!ht]
\centering
\includegraphics[width=.92\textwidth]{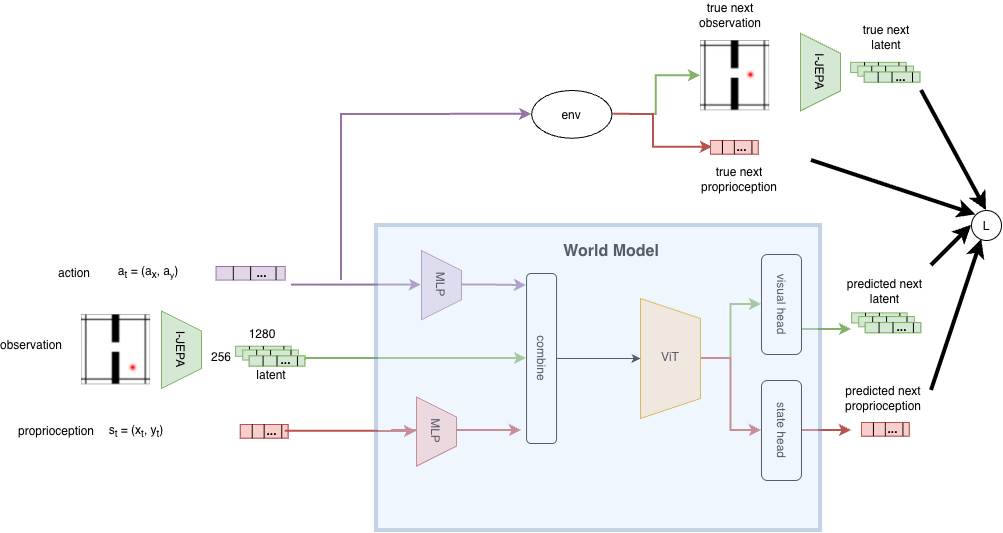}
\caption{Detailed world-model training architecture. A frozen I-JEPA encoder maps the current and next observations to visual embeddings. The model combines the current embeddings and proprioceptive state with a macro-action comprising five successive 2D actions. An action-conditioned transformer predicts the next visual embeddings and proprioceptive state; both predictions are trained against their observed targets using MSE.}
\label{fig:wm-training-detailed}
\end{figure}

Figure~\ref{fig:wm-training-detailed} shows the world-model training architecture. A frozen, pretrained I-JEPA ViT-H/14 encoder maps each $224\times224$ observation to a grid of $16\times16$ latent embeddings, each with 1280 dimensions. The model also receives the ball's proprioceptive state: position $(x,y)$ in DotWall, and position and velocity $(x,y,v_x,v_y)$ in PointMaze.

One world-model step corresponds to five successive two-dimensional primitive actions. We concatenate these actions into a single 10-dimensional \emph{macro-action}. Separate MLPs embed the macro-action and proprioceptive state; their outputs are combined with each visual embedding and passed through a transformer with spatial and temporal position embeddings. A visual head predicts the next grid of embeddings, while a proprioceptive head predicts the state after the five actions have been executed.

For training, the frozen I-JEPA encoder supplies the target embeddings from the observed next frame. We minimize the MSE between predicted and target visual embeddings, together with a weighted MSE between the predicted and observed next proprioceptive states. The proprioceptive states are normalized for this loss. The training target is therefore the observed outcome of each macro-action; no goal distance or geodesic target is used to train the world model.

\appsubsection{Environments and dataset generation}
\label{app:datasets}

We study two image-goal navigation environments (Fig.~\ref{fig:environments}). In \textbf{DotWall}, a ball moves between two rooms separated by a wall with a doorway. The world model used for the main planning experiments was trained on a fixed layout, with the wall at $x=32$ and the doorway at $y=30$. In \textbf{PointMaze}, the ball must navigate a U-shaped corridor, sometimes travelling away from the goal before it can approach it along a legal route. DotWall provides the ball position $(x,y)$ as proprioception; PointMaze provides position and velocity $(x,y,v_x,v_y)$.

We generated a separate dataset of 2,000 simulator episodes for each world model, split into 1,600 training, 200 validation, and 200 test episodes. DotWall episodes contain 50 primitive actions. Their actions follow a random policy with correlated headings and varying action lengths, producing motion and wall-contact examples without prescribing a route to a goal. PointMaze episodes contain 100 primitive actions sampled from the environment's action bounds. In the PointMaze setting used here, each primitive action advances five physics steps. We record rendered $224\times224$ RGB observations, actions, and proprioceptive states. The frozen I-JEPA encoder supplies the visual embeddings; five successive 2D actions are concatenated into each 10-dimensional world-model macro-action.

The \emph{adapter-fitting experience} is collected separately, after world-model training. For DotWall, planning rollouts start in progressively wider bands around 12 goals on the fixed layout. PointMaze uses the same collection principle around four training goals. Successful and failed rollouts supply the ranking comparisons described in the main text; the held-out goals used for paired planning evaluation are excluded from adapter fitting.

\appsubsection{Ranking observations from failed episodes}
\label{app:failed-episodes}


A failed episode need not be devoid of progress: it may approach a route that has previously reached the same goal before moving away. We sample up to five distinct observations along each failed episode and compare each with the successful routes stored for that goal. For a sampled observation $f$ and successful route $R_k(g)$, we find the observation on that route with the lowest latent cost to $f$ under the adapter frozen for the current collection round. We then average the three lowest such costs from \emph{different} successful routes. Thus, a failed observation that is close in latent space to several successful routes receives a lower score and is ranked ahead of one that is far from those routes. Near-ties and goals with too few successful reference routes are excluded from this source of pairs.

\begin{figure}[!ht]
\centering
\includegraphics[width=.92\textwidth]{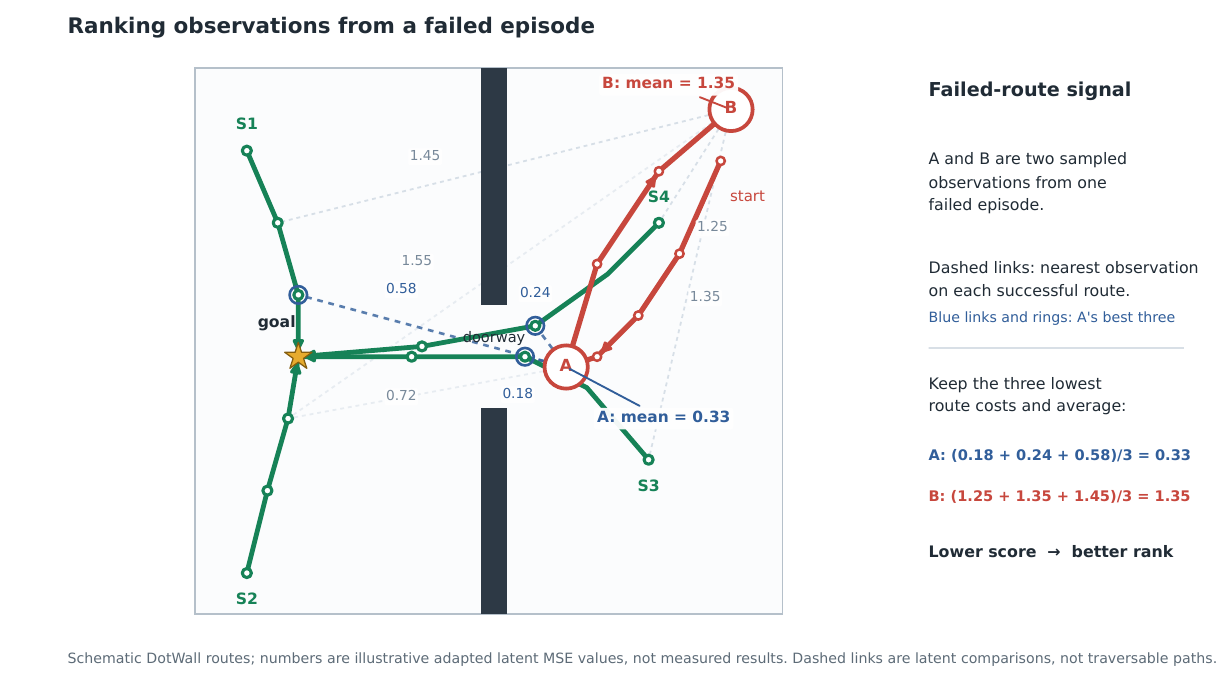}
\caption{Schematic of the failed-episode signal in DotWall. Green routes $S_1$--$S_4$ reach the goal; the red route fails. Dots are visited observations and circles mark two sampled failed observations. Dashed links represent latent comparisons with each successful route, not traversable paths. Highlighted blue links are the three lowest costs for $A$. All displayed values are illustrative.}
\label{fig:failed-signal}
\end{figure}

Figure~\ref{fig:failed-signal} illustrates the calculation with two sampled observations, $A$ and $B$, on a single failed route. For each successful route, a dashed line reaches the nearest observation in latent cost, rather than necessarily the nearest one in physical distance. The three lowest route costs are highlighted for $A$; their average is 0.33, compared with 1.35 for $B$. We therefore form the weak preference $A \prec B$, meaning that $A$ should receive a lower adapted latent cost to the goal. The numbers in the schematic are illustrative, not experimental measurements. This rule uses successful experience as a reference without supplying geodesic progress labels to the adapter.

\end{document}